\documentclass[letterpaper]{article}
\usepackage{uai2020}
\usepackage[margin=1in]{geometry}

\usepackage{times}

\title{On the Relationship Between Probabilistic Circuits and\\Determinantal Point Processes}

\author{ \textbf{Honghua Zhang}
  \and \textbf{Steven Holtzen}
  \and \textbf{Guy Van den Broeck} \\
  Computer Science Department \\ University of California, Los Angeles \\
  \texttt{\{hzhang19,sholtzen,guyvdb\}@cs.ucla.edu}
}

\usepackage{subcaption}
\usepackage{natbib}
\usepackage{latexsym}
\usepackage{amsmath}
\usepackage{amssymb}
\usepackage{amsthm}
\usepackage{epsfig}
\usepackage{graphicx}
\usepackage{mathrsfs}
\usepackage{blkarray}
\usepackage{enumitem}
\usepackage{tikz}
\usetikzlibrary{shapes.misc, positioning, shapes.geometric, backgrounds,
  patterns, fit, calc}

\tikzstyle{circuit}=[
  >=latex, thick, >=stealth, font=\small,auto,scale=0.9,every node/.style={scale=0.9}
]
\tikzstyle{internal}=[
  inner sep=0pt,fill=white
]

\tikzstyle{edge}=[
    line width=0.9
]
\tikzstyle{terminal}=[
  draw,inner sep=2.5pt, font=\small,fill=white
]

\newcommand{\R}{\mathbb{R}}
\newcommand{\Q}{\mathbb{Q}}

\newcommand{\N}{\mathbb{N}}

\newcommand{\td}{\widetilde}
\newcommand{\bigO}{\mathcal{O}}
\newcommand{\op}{\operatorname}
\newcommand{\ol}{\overline}
\newcommand{\mbf}{\mathbf}
\newcommand{\mcal}{\mathcal}
\newcommand{\mscr}{\mathscr}
\newcommand{\disp}{\displaystyle}
\newcommand\given[1][]{\:#1\vert\:}
\newcommand\widebar[1]{\mathop{\overline{#1}}}
\newcommand{\eval}{\bigr\rvert}

\theoremstyle{plain} 
\newtheorem{thm}{Theorem}

\newtheorem{lem}{Lemma}
\newtheorem{prop}{Proposition}

\theoremstyle{definition}

\newtheorem{defn}{Definition}

\theoremstyle{remark}

\usepackage{microtype}
\usepackage{graphicx}
\usepackage{booktabs} 

\usepackage{hyperref}

\begin{document}
\maketitle

\begin{abstract}
Scaling probabilistic models to large realistic problems and datasets is a
key challenge in machine learning. Central to this effort is
the development of \emph{tractable probabilistic models} (TPMs):
models whose structure guarantees efficient probabilistic inference
algorithms. The current landscape of TPMs is fragmented: there
exist various kinds of TPMs with different strengths and weaknesses. 
Two of the most prominent classes of TPMs are determinantal point processes (DPPs) and probabilistic circuits (PCs). 
This paper provides the first systematic study of their relationship. 
We propose a unified analysis and shared language for discussing DPPs and PCs.
Then we establish theoretical barriers for the unification of these two families,  
and prove that there are cases where DPPs have no compact representation as a class of PCs. 
We close with a perspective on the central problem of unifying these tractable models.
\end{abstract}

\section{INTRODUCTION}

Probabilistic modeling has become a central area of study in machine learning.
The key challenge in applying
probabilistic modeling is scaling to large datasets and models: in many cases,
probabilistic inference quickly becomes intractable as the models grow in size
and sophistication, and in general the task is \#P-hard~\citep{roth1996hardness}.
The field of \emph{tractable probabilistic modeling} (TPM) seeks to identify
classes of probabilistic models that (1) guarantee efficient probabilistic
reasoning, and (2) can compactly represent interesting rich probability
distributions. 

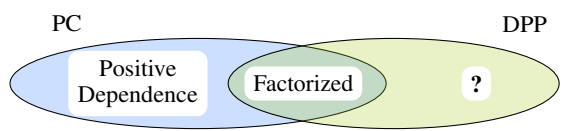
\begin{figure}
\hspace{-8pt}
\begin{tikzpicture}[
  every path/.style = {
   ->,
   > = stealth, 
   rounded corners},
  state/.style = {
    fill = white,
    text centered
  },
  node distance=1cm]

  \definecolor{color1}{HTML}{E7AD00}
  \definecolor{color2}{HTML}{A5CC19}
  \definecolor{color3}{HTML}{33B29A}
  \definecolor{color4}{HTML}{3380FF}
  \definecolor{color5}{HTML}{9033FF}
  \definecolor{color6}{HTML}{E5003D}

\begin{scope}[
  opacity = 1,
  fill opacity = 0.25,
  text opacity = 1,
  text width = 6em,
  text centered]

  \draw [fill = color4] (-0.2,0) ellipse (2.5 and 0.6)
    node [above left = 0.8] {\small PC};
  \draw [fill = color2] (2.4,0) ellipse (2.2 and 0.6)
    node [above right = 0.8] {\small DPP};

\end{scope}

\begin{scope}
   \tikzset{font=\footnotesize}
  \node[state] at (3.5,0) {\textbf{?}};
  \node[state] at (1.2,0) {Factorized};
  \node[state, text width=46] at (-1.0,0) {Positive\\Dependence};

\end{scope}

\end{tikzpicture}
\caption{Known relationships between probabilistic circuits (PC) and determinantal point processes (DPP).}
\label{fig:venn_diagram_simple}
\vspace{-0.5cm}
\end{figure}

In recent years there has been a proliferation of development of different
classes of TPMs called tractable \emph{probabilistic
circuits}~\citep{AAAI-Tutorial, LecNoAAAI20}. Each strikes a different balance
between restrictions on the representation and increasing ease of learning and
answering queries. Examples include bounded-treewidth graphical
models~\citep{meila2000learning}, sum-product networks (SPNs)
\citep{poon2011sum, peharz2019random}, probabilistic sentential decision
diagrams (PSDDs)~\citep{kisa2014probabilistic}, arithmetic
circuits~\citep{darwiche2009modeling}, and cutset
networks~\citep{rahman2016learning}.

In a separate line of research, a TPM called \emph{determinantal point processes}
(DPPs) has been the topic of intense investigation, in particular because of their wide-ranging applications in machine learning~\citep{MAL-044, borodin2009determinantal, krause2005note}.
They excel at representing certain types of distributions, but do so in ways
that are distinct from how probabilistic circuits work.
This raises the central questions of this paper:
\emph{Are DPPs and PCs really distinct in their ability to efficiently represent and reason about probability distributions}?
Moreover, if we understand their relationship, 
\emph{to what extent can their strengths and weaknesses be unified into one general TPM}? More broadly, one wonders, what is the essence of tractable probabilistic~modeling? 

Figure~\ref{fig:venn_diagram_simple} summarizes what is known about this relationship. PCs can represent positive dependencies beyond the reach of DPPs, and simple factorized distributions can be represented in both TPMs. Our key contribution is to fill in the ``?'' that represents a gap in the literature: it is currently unknown whether circuits can represent DPPs.

Section~\ref{sec:motiv} formally introduces DPPs and PCs and further motivates our research. Section~\ref{sec:unify} discusses the relative
strengths and weaknesses of DPPs and PCs as tractable families, and gives a semantic foundation for unifying them. Section~\ref{sec:reprdpp} poses the problem of representing DPPs as circuits,
and proves that such a representation is always inefficient. Finally, Section~\ref{sec:discussion} outlines perspectives and future directions for 
unifying these two families.

\section{BACKGROUND AND MOTIVATION}
\label{sec:motiv}

This paper studies probabilistic models that are representations of discrete probability distributions $\Pr(X_1,X_2,\dots)$, where $X$ denotes a binary random variable. Assignments of values to the random variables are written~$X\!=\!x$. Sets of binary random variables and their joint assignments are written in bold (e.g.,~$\mathbf{X}\!=\!\mathbf{x}$).

Our discussion focuses on two defining characteristics of probabilistic
models: their \emph{expressive efficiency}~\citep{martens2014expressive}
and \emph{tractability}. We also refer to expressive efficiency as \emph{succinctness} for short.
A probabilistic model is
efficient in terms of expressiveness (or succinct) for a class of distributions if it can compactly represent those
distributions -- i.e., the size of the model (for some appropriate definition of
size) is polynomial in the number of random
variables. A \emph{query} -- for instance, computing the marginal probability of an
event -- is \emph{tractable} for a model if it can be computed in
time polynomial in the size of the model.

In particular, we consider two well-known tractable probabilistic models that achieve their
tractability in strikingly different ways: determinantal point processes (DPPs)
and probabilistic circuits (PCs). Next, we formally introduce
the semantics of their representation, and compare how they achieve their tractability, motivating the key research questions that this paper seeks to answer.

\subsection{PROBABILISTIC REPRESENTATIONS}
This section briefly describes DPPs and PCs through a unified notation and vocabulary.

\subsubsection{DPPs as L-ensembles}
\label{sec:dpp_as_l}
Within machine learning, DPPs are most often represented by means of an 
\emph{L-ensemble}~\citep{borodin2005eynard}:\footnote{Although not every DPP 
is an L-ensemble, \cite{MAL-044} show that DPPs that assign non-zero probability 
to the empty set (the all-false assignment) are L-ensembles. Hence, this is a weak assumption in most applications.}

\begin{defn}
  \label{def:lensemble}
A probability distribution $\Pr$ over $n$ binary random variables $\mbf{X} = (X_1, \cdots, X_n)$ is an \emph{L-ensemble} if there exists a symmetric
positive semidefinite matrix $L \in \R^{n \times n}$ such that for all
$\mbf{x} = (x_1, \cdots, x_n) \in \{0, 1\}^{n}$,
\begin{align}
\Pr(\mbf{X} = \mbf{x}) \propto \det(L_{\mbf{x}}),
  \label{eq:dppkernel}
\end{align}
where $L_{\mbf{x}} = [L_{ij}]_{x_{i} = 1, x_{j} = 1}$ denotes the submatrix of $L$
indexed by those $i,j$ where $x_{i} = 1$ and $x_{j} = 1$. The matrix $L$ is called the
\emph{kernel} for the L-ensemble.
\end{defn}
To ensure that the distribution sums to one, it is necessary to divide Equation~\ref{eq:dppkernel} by $\det(L + I)$, where $I$ is the $n \times n$ identity matrix \citep{MAL-044}.

A first barrier for linking DPPs and other probabilistic modeling frameworks
is notational.
While Definition~\ref{def:lensemble} characterizes DPPs as distributions over $n$ binary random variables $\mathbf{X}$, which is typical in the probabilistic graphical model literature, the DPP literature instead prefers to characterize them as distributions over \emph{sets}. Fortunately there is a simple mapping between these interpretations.

Formally, given the finite {\it ground set} $\mcal{Y} = \{1, \cdots, n\}$, a DPP
assigns a probability to each subset of $\mcal{Y}$.
The binary random variables $\mbf{X}$ define the random subset
$\mathbf{Y} = \{i \in \mcal{Y}: X_i = 1\}$; that is, we can view variable $X_i$ as the indicator variable for item $i \in \mcal{Y}$ being in the random set~$\mbf{Y}$.
Similarly, each set assignment $\mathbf{Y}\!=\!A$ where $A \subset \mcal{Y}$ corresponds to a binary assignment $\mbf{X}\!=\!\mbf{x}_{A}$ that sets $((X_i = 1)_{i \in A}, (X_j = 0)_{j \notin A})$.

Then, for any set value $A \subset \mcal{Y}$, we can write
$$\Pr(\mbf{Y} = A) = \Pr({\mbf{X} = \mbf{x}_{A}}) \propto \det(L_{\mbf{x}_A}) = \det(L_{A}),$$
where $L_{A} = [L_{ij}]_{i, j \in A}$ is the submatrix of $L$ indexed by elements
in $A$. For simplicity, we will denote $\mcal{Y} - A$ by $\widebar{A}$. Viewing
a distribution from the perspective of subsets can be more intuitive for describing certain properties of DPPs -- we will use both perspectives interchangeably.

Consider the following example of an L-ensemble defined over variables $X_1$, $X_2$, and $X_3$:
\begin{align*}
 L =
\begin{blockarray}{cccc}
X_1 & X_2 & X_3 & \\
\begin{block}{[ccc]c}
  2 & 1.1 & 1.4 & X_1 \\
  1.1 & 2.5 & 0.5 & X_2 \\
  1.4 & 0.5 & 3 & X_3 \\
\end{block}
\end{blockarray}
\end{align*}
Then, the probability of the assignment
$\mbf{X} = (1, 0, 1)$, or equivalently, subset $\mbf{Y} = \{1, 3\}$ is given by:
\begin{align*}
  &\Pr(\mbf{X} = (1, 0, 1)) = \Pr(\mbf{Y} = \{1, 3\})\\
&\quad= \frac{\det(L_{\{1, 3\}})}{\det(L + I)} = \frac{1}{31.09}
\begin{vmatrix}
 2 &  1.4 \\
 1.4 &  3 \\
\end{vmatrix} = 0.13.
\end{align*}

L-ensembles are clearly succinct: they are specified by $\bigO(n^2)$ parameters in kernel $L$.
Moreover, evaluating the probability of an instantiation is efficient: it simply requires computing the determinant of a
submatrix of kernel $L$, which takes $\bigO(n^3)$ time using Gaussian elimination.

One of the most important properties of DPPs is that they represent \emph{global negative dependence}, which 
is discussed in detail in Section~\ref{sec:unify}. This property also brings a key limitation to DPPs: they
cannot represent distributions with \emph{any} positive dependence.
Specifically, for a DPP, it must always be
the case that
$\Pr(X_i = 1, X_j = 1) \le \Pr(X_i = 1)\Pr(X_j = 1)$. In Figure~\ref{fig:venn_diagram_simple},
probability distributions with positive dependence lie in the blue section.

\subsubsection{Probabilistic Circuits}
Next, we introduce the basics of probabilistic circuits~\citep{AAAI-Tutorial, LecNoAAAI20} -- a class of probabilistic models that is strikingly different from L-ensembles. 

\begin{figure}
 \centering
     \begin{subfigure}[b]{0.45\linewidth}
       \centering
       \includegraphics[width=\textwidth]{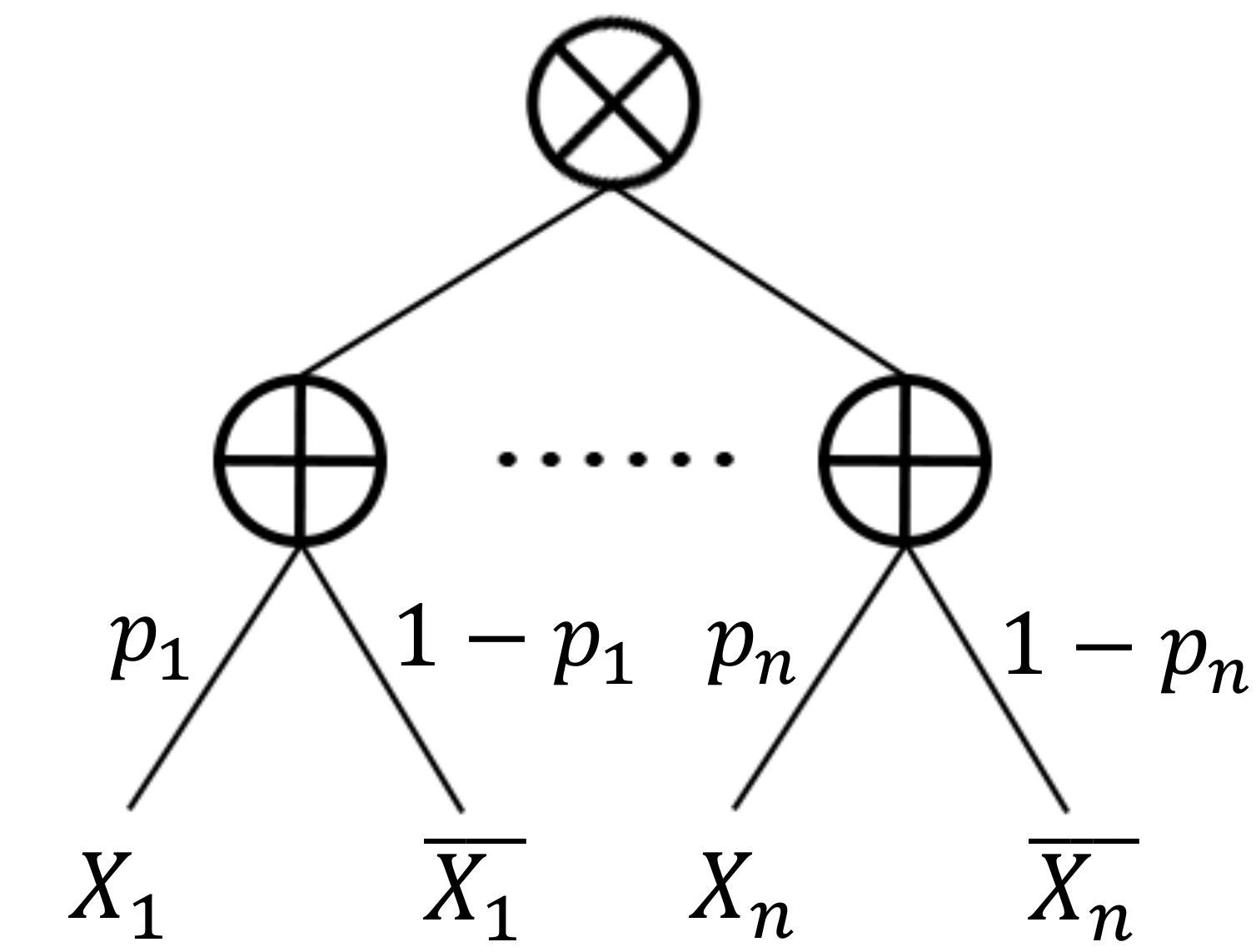}
       \caption{A deterministic and decomposable PC.}
       \label{fig:berncircuit}
     \end{subfigure}
     ~
     \begin{subfigure}[b]{0.45\linewidth}
       \centering
       \begin{align*}
         \begin{bmatrix}
           \frac{p_1}{1 - p_1} &  &  & 0\\
               & \frac{p_2}{1 - p_2} &  &   \\
               &  & \ddots &  \\
               0&     & & \frac{p_n}{1 - p_n}
         \end{bmatrix}
       \end{align*}
       \caption{The kernel matrix $L$ for Figure~\ref{fig:berncircuit}.}
       \label{fig:bernkern}
     \end{subfigure}
    \\ 
     \begin{subfigure}[b]{0.45\linewidth}
      \centering
      \includegraphics[width=\textwidth]{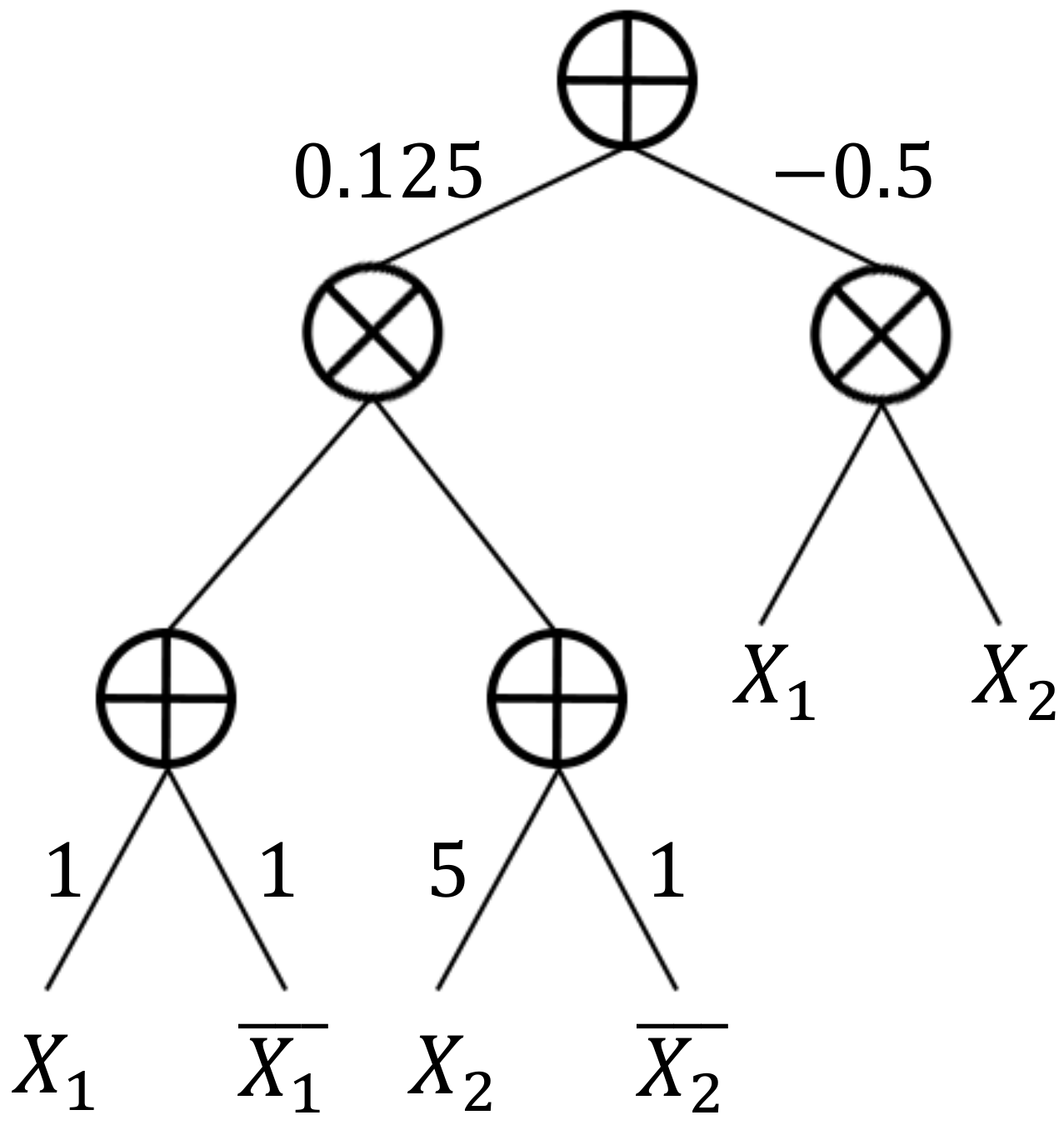}
      \caption{A decomposable PC for the L-ensemble in Figure~\ref{fig:3elemdpp}.}
     \label{fig:3elemcirc}
     \end{subfigure}
     ~
     \begin{subfigure}[b]{0.45\linewidth}
        \begin{align*}
         L &= \begin{blockarray}{ccc}
                X_1 & X_2 & \\
                \begin{block}{[cc]c}
                1 & 2 & X_1\\
                2 & 5 & X_2\\
                \end{block}
              \end{blockarray}
        \\
         K &= \begin{blockarray}{ccc}
                X_1 & X_2 & \\
                \begin{block}{[cc]c}
                0.25 & 0.25 & X_1\\
                0.25 & 0.75 & X_2\\
                \end{block}
              \end{blockarray}
       \end{align*}
       \caption{The kernel $L$ and marginal kernel $K$.}
       \label{fig:3elemdpp}
     \end{subfigure}
     \caption{Motivating DPPs and their equivalent PCs.}
    \label{fig:motivation}
\end{figure}

\begin{defn}
A \emph{probabilistic circuit} (PC) is a directed acyclic graph consisting of three types of
nodes:
\vspace{-0.3cm}
\begin{enumerate}[noitemsep,leftmargin=*]
\item \emph{Sums} $\bigoplus$ with weighted edges to their children;
\item \emph{Products} $\bigotimes$ with unweighted edges to their children;
\item \emph{Variable leaves}, here assumed to be $X_i$ or~$\widebar{X_i}$.
\end{enumerate}
\vspace{-0.1cm}

For a given assignment $\mbf{X}\!=\!\mbf{x}$, the probabilistic circuit~$\mscr{A}$ \emph{evaluates} to a number $\mscr{A}(\mbf{x})$, which is obtained by (i)~replacing variable leaves $X_i$ by $x_i$, (ii)~replacing variable leaves $\widebar{X_i}$ by $1-x_i$, (iii)~evaluating product nodes as taking the product over their children, and (iv)~evaluating sum nodes as taking a weighted sum over their children.

Finally, a probabilistic circuit $\mscr{A}$ with variable leaves $\mbf{X} = (X_1, \cdots, X_n)$ represents the probability distribution
\begin{align}
 \Pr(\mbf{X}\!=\!\mbf{x}) \propto \mscr{A}(\mbf{x}).   \label{eq:circuit_semantics}
\end{align}
\end{defn}
Hence, evaluating the probability of an instantiation for a PC is simply a bottom-up pass, which is a linear-time algorithm in the size of the PC.
As with DPPs, the semantics involve a normalizing constant. Section~\ref{s:tractability_of_pcs} will discuss how to compute this normalization efficiently by means of tractable marginalization.

A PC's edge weights are called its \emph{parameters}, which are usually 
assumed to be non-negative. We will relax this assumption 
in this paper, and also consider circuits with possibly negative parameters.
Note that a PC with negative parameters does not necessarily represent a
probability distribution.

An example PC that represents a fully-factorized distribution is given in
Figure~\ref{fig:berncircuit}. This circuit evaluates to $$\Pr(\mbf{X}\!=\!\mbf{x}) = \mscr{A}(\mbf{x}) = \prod_i \left(x_i p_i + (1-x_i) (1-p_i) \right).$$
This fully-factorized distribution is also an L-ensemble, and the kernel $L$ for which is shown in Figure~\ref{fig:bernkern}.
Hence, we know that the intersection in Figure~\ref{fig:venn_diagram_simple} is non-empty.

\subsubsection{Motivating Questions}
The prior sections outlined the semantics of DPPs and PCs for computing the probability of an instantiation of random variables. 
However, these representations are currently fundamentally distinct.
A PC is a computation graph: it directly describes how to carry out the computation of an instantiation probability.
A DPP, on the other hand, leaves unspecified how to compute the determinant.
This hints at the following research questions:
\begin{description}[noitemsep, topsep=0pt]
\item[(Q1)] Can an algorithm that computes determinants be efficiently captured by a computation graph that is a probabilistic circuit?
\item[(Q2)] Or, do there exist DPPs that PCs are not expressive enough to represent succinctly?
\end{description}

We saw in the previous section that DPPs cannot represent distributions
with positive dependence. In contrast, PCs are able 
to represent \emph{any} probability distributions over binary random variables, 
including the ones with positive dependence. Indeed, we can always trivially construct a PC with a large $\bigoplus$ root node that enumerates all $2^{n}$ instantiations with their probabilities as edge weights.
The real question is, however, can PCs represents DPPs efficiently? 
To answer this question, we first need to understand the requirements for tractable inference.

\subsection{TRACTABLE PROBABILISTIC INFERENCE} \label{s:tractable_inference}
The previous section defined L-ensembles and PCs, and showed that evaluating the probability of complete assignments is tractable in both models.
However, the purpose of tractable probabilistic models is to answer more interesting queries efficiently.
This section briefly discusses the probabilistic inference tasks that can be performed tractably for DPPs and PCs; in particular,
we will focus on two tasks: computing marginals and MAP inference.

\subsubsection{Tractability of DPPs}

As an alternative to L-ensemble kernels, DPPs can more generally be represented by their \emph{marginal kernels}.
\begin{defn}
A probability distribution $\Pr$ over $n$ binary random variables $\mbf{X} = (X_1, \cdots, X_n)$ is a {\it determinantal point process} if
there exists a symmetric positive semidefinite matrix $K \in \R^{n \times n}$ such
that for all $A \subset \{1, \cdots, n\}$,
\begin{align}
\Pr((X_i \!=\! 1)_{i \in A}) = \det(K_{\mbf{x}_{A}}),  \label{eq:dpp_marginal_positive}
\end{align}
or, equivalently, from the perspective of subsets,
$$\Pr(A \subset \mbf{Y}) = \det(K_A).$$
The matrix $K$ is called the \emph{marginal kernel} for the DPP.
\end{defn}
Note that we are computing a marginal probability: variables $X_i$ where $i \not\in A$ are marginalized out. From the set perspective, we are asking whether the random subset contains the elements of $A$, not whether it is equal to $A$.

One can use a generalized version of Equation~\ref{eq:dpp_marginal_positive} to compute
the general marginals $\Pr((X_i \!=\! 1)_{i \in A}, (X_j \!=\! 0)_{j \in B})$ efficiently, where $A, B \subseteq \{1, \cdots, n\}$. We refer to \citet{MAL-044} for further details.

A seminal result in the DPP literature is that every L-ensemble with kernel
$L$ is a DPP with marginal kernel $K = L(L + I)^{-1}$~\citep{macchi1975coincidence}.
In light of this, it is clear that computing marginals is efficient for
L-ensembles in general, because computing the inverse and determinant can both be done in polynomial time.
However, MAP inference, where the task is to find the most likely world given evidence, is known to be NP-hard for L-ensembles~\citep{ko1995exact}.

\subsubsection{Tractability of PCs} \label{s:tractability_of_pcs}

Next we discuss the same queries for PCs -- we will show they are tractable for entirely different, structural reasons.

For an arbitrary probabilistic circuit, most
probabilistic inference tasks, including marginals and MAP inference,
are computationally hard in the circuit size and therefore
inefficient. In order to guarantee the efficient evaluation of queries it is
therefore necessary to impose further constraints on the structure of the
circuit. There are two well-known structural properties of probabilistic
circuits~\citep{darwiche2002knowledge,LecNoAAAI20}:
\begin{defn}
For a probabilistic circuit,
\begin{enumerate}[noitemsep]
\item A $\bigotimes$ node is \emph{decomposable} if its inputs depend
  on disjoint sets of variable nodes.
\item A $\bigoplus$  node is \emph{deterministic} if at most one of its inputs
  can be non-zero for any assignment to the variable leaf nodes.
\end{enumerate}
We say a PC is decomposable if all of its $\bigotimes$ nodes are decomposable;
a PC is deterministic if all of its $\bigoplus$ nodes are deterministic.
\end{defn}
In this paper, we will also consider a special class of PCs:
Probabilistic Sentential Decision Diagrams (PSDDs)~\citep{kisa2014probabilistic}. At a high-level,
a PSDD is a deterministic \emph{and} decomposable PC with some additional structural
properties. Because of these structural properties, PSDDs
come with additional tractability guarantees~\citep{KhosraviNeurIPS19} and strong local properties, which will be explained later; for further details
please refer to~\citet{kisa2014probabilistic}.

Let $\mscr{A}$ be a PC over $X_1, \cdots, X_n$. If $\mscr{A}$ is decomposable,
then we can efficiently compute its marginals: for disjoint
$A, B \subset \{1, \cdots, n\}$, the marginal probability
$\Pr((X_i = 1)_{i \in A}, (X_j = 0)_{j \in B})$ is given by the evaluation of
$\mscr{A}$ with the following input:
$$
\begin{cases}
X_{i} = 1, \widebar{X_{i}} = 0 & \text{ if } i \in A \\
X_{i} = 0, \widebar{X_{i}} = 1 & \text{ if } i \in B \\
X_{i} = 1, \widebar{X_{i}} = 1 & \text{otherwise.}
\end{cases}
$$
Thus, for a decomposable PC, the time complexity
of marginal computation is linear in the size of the circuit.

Though computing marginals is tractable
for both L-ensembles and decomposable PCs, it is done very differently.
For a decomposable PC, we only need to
evaluate the circuit with respect to certain input configurations.
That is to say, if we can represent an L-ensemble with a
decomposable PC, we would have a unified representation for both the kernel
$L$ and marginal kernel $K$ of an L-ensemble.
For example, Figure~\ref{fig:3elemcirc} shows a decomposable PC that represents 
the L-ensemble in Figure \ref{fig:3elemdpp}; say we want to compute the marginal 
distribution $\Pr(X_2 = 1)$: for the PC, we plug in $X_1 = 1, \widebar{X_1} = 1,
X_2 = 1, \widebar{X_2} = 0$ and it evaluates to $0.75$, which corresponds to
the value of the entry $K_{22} = \Pr(2 \in \mbf{Y})$. 

For a deterministic \emph{and} decomposable PC, MAP inference can be done in time polynomial in the size of the PC, which implies that MAP inference is also tractable for PSDDs.
When a PC is unnormalized, we also need to compute its normalizing constant, the time complexity of which
is linear in the size of a decomposable PC.
 
\section{A UNIFIED TRACTABLE MODEL}
\label{sec:unify}
A unique property of DPPs is that they are
\emph{tractable} representations of probability distributions that express
{\it global negative dependence}.
Graphical models are limited
by their local nature thus cannot effectively model such global negative dependence.
For example, consider a DPP that assigns non-zero probabilities to all \emph{proper}
subsets of $\{1, \cdots, n\}$ and zero probability to the whole set:
a Markov Random Field that is not the complete graph (or a factor graph
without a factor that connects all nodes), cannot model this distribution.
To further demonstrate what global negative dependence means
we consider the geometric interpretation of determinants \citep{MAL-044}.

Consider an L-ensemble $P_{L}$. Since kernel $L$
is positive semidefinite, by the Spectral Theorem, there exists a matrix $B$
such that $L = B^{T}B$. We denote the columns of $B$ by $B_{i}$, where $1 \leq i \leq n$.
Then for $A \subset \mathcal{Y}$,
$$P_{L}(\mbf{Y} = A) \propto \det(L_A) = \text{Vol}^{2}(\{B_i\}_{i \in A})$$
where $\text{Vol}^{2}(\{B_i\}_{i \in A})$ is the squared $|A|$-dimensional
volume of the parallelepiped spanned by $\{B_i\}_{i \in A}$. Intuitively, we can
view $B_{i}$ as the feature vector for element $i \in \mathcal{Y}$.
If the elements in a set $A \subset \mathcal{Y}$
are very similar to each other, 
then the angles between their feature vectors are small;
then, the volume spanned by their feature vectors are small, which implies that
the probability of $A$ (and any set that contains $A$) is small. From this perspective,
we can see that DPPs always encourage {\it diversity} by assigning higher probabilities
to sets of elements that are more different from each other.
The ability to tractably model diversity makes DPP a useful class of TPMs
in many applications~\citep{mariet2016kronecker}, such as document and video
summarization~\citep{chao2015large, lin2012learning}, recommender
systems~\citep{zhou2010solving}, and object retrieval~\citep{affandi2014learning}.

Despite being a tractable model that expresses diversity and negative dependencies,
DPP is not very flexible in the sense it cannot model {\it any} positive correlations.
Hence, there arises the question whether we can tractably model such diversity
by a more flexible TPM that allows some positive correlations.
Thus, we consider the problem of tractably representing
DPPs (or L-ensembles) by PCs, which are much more flexible
than DPPs.

\subsection{DPPs AS PCs}
In this section we discuss some potential solutions to the problem of
representing L-ensembles by PCs; we start by introducing the \emph{symbolic kernel} for an L-ensemble.
\begin{defn}
Let $L \in \R^{n \times n}$ be the kernel for an L-ensemble over $X_1, \dots, X_n$;
the corresponding \emph{symbolic kernel} $L^{*}$ is given by:
$$
{L^{*}}_{ij} = 
\begin{cases}
    L_{ij}X_i + \widebar{X_i} & \text{if } i = j \\
    L_{ij}X_iX_j & \text{if } i \neq j
\end{cases}
$$
With the symbolic kernel $L^{*}$, probabilities of an L-ensemble can be written in a different way:
$$
\textstyle{
\Pr(\mbf{X} = \mbf{x}) = \det(L + I)^{-1} ( \det(L^{*})\eval_{\mbf{X} = \mbf{x}} )},$$
where the notation $\eval_{\mbf{X} = \mbf{x}}$ replaces the symbolic variables
with assignments.
\end{defn}
To get a concrete understanding of the symbolic kernel, we consider the example L-ensemble presented in 
Section~\ref{sec:dpp_as_l}; the corresponding symbolic kernel is
$$L^{*} =
\begin{bmatrix}
 2X_{1} + \widebar{X_{1}} & 1.1X_{1}X_{2} & 1.4X_{1}X_{3} \\
 1.1X_{1}X_{2} & 2.5X_{2} + \widebar{X_{2}} & 0.5X_{2}X_{3} \\
 1.4X_{1}X_{3} & 0.5X_{2}X_{3} & 3X_{3} + \widebar{X_{3}} \\
\end{bmatrix}.
$$
Then the probability of the instantiation $(1, 0, 1)$ is 
\begin{align*}
  &\Pr(X_1 = 1, X_2 = 0, X_3 = 1) \\
= &{\det(L + I)}^{-1}\det(L^{*})\eval_{X_1 = 1, X_2 = 0, X_3 = 1} \\
= &\frac{1}{\det(L + I)}
    \begin{vmatrix}
        2 & 0 & 1.4 \\
        0 & 1 & 0 \\
        1.4 & 0 & 3
    \end{vmatrix} 
= \frac{1}{\det(L + I)}
    \begin{vmatrix}
        2 & 1.4 \\
        1.4 & 3
    \end{vmatrix}
\end{align*}
Note that each entry of the symbolic kernel $L^{*}$ can be easily represented by a PC;
hence, with the symbolic kernel, the problem of representing DPPs by
polynomial-size PCs can be reduced to the problem of representing the determinant
function (with entries of the matrix being the inputs) by a polynomial-size PC.

Our first observation is that a PC that computes the determinant function must
have at least one {\it negative} parameter, which is a non-standard assumption
for PCs: a PC with non-negative parameters would always output
non-negative numbers for matrices with all positive
entries, which, however, could have negative determinant.

Our second observation is that constructing a polynomial-size
PC that computes the determinant is not easy.
Gaussian elimination, for instance, would not work here as it requires branching and
division. The Laplace expansion of the determinant does not require branching or
division, but it has to compute the determinants of exponentially many submatrices.
Nonetheless, a combinatorial algorithm proposed by \citet{mahajan1997combinatorial}
gives us a polynomial-size PC for the determinant; unfortunately we observe that
the induced PC that represents DPPs is neither deterministic nor decomposable.

Of course, one might say that we \emph{do not have to} reduce the problem
of representing DPPs by PCs to the problem of constructing a PC that
computes the determinant. We will show in the following section that, 
roughly speaking, polynomial-size PCs that represents DPPs cannot be
deterministic or decomposable.

\section{BARRIERS TO REPRESENTING DPPs WITH PCs}
\label{sec:reprdpp}
In this section, we prove that in general,
DPPs cannot be tractably represented by certain subclasses of PCs.
In particular, we consider three different classes of PCs:
PSDDs, deterministic PCs and decomposable PCs.
By ``tractable representation,''
we mean PCs of polynomial size with respect to the number of
input variables.

\subsection{PSDDs CANNOT TRACTABLY REPRESENT ALMOST ALL L-ENSEMBLEs}
Though we will show in the following subsections that neither
deterministic PCs nor decomposable PCs can tractably represent DPPs
in general, for PSDDs, we prove a much stronger result.
Instead of constructing a small subclass of DPPs that cannot
be tractably represented by PSDDs,
we prove that {\it almost all} L-ensembles have no polynomial-size
PSDD representations, which we formalize as the following theorem.
\begin{thm} \label{thm:PSDD}
Let $n \in \N^{+}$, $M > 0$;
let $B$ be a matrix drawn from the uniform distribution over
$[-M, M]^{n \times n}$. Let $L = B^{T}B$. Then with probability $1$,
the L-ensemble with kernel $L$ cannot be tractably represented by
any PSDD.
\end{thm}
The proof for Theorem~\ref{thm:PSDD} relies on \emph{algebraically independent
numbers}:
\begin{defn}
Let $F$ be a field; define
$F[t_{1}, \ldots, t_{k}]$ to be the set of polynomials over {\it indeterminates}
$t_{1}, \cdots, t_{k}$ with coefficients in $F$. We call $0$ the \emph{trivial polynomial}.
\end{defn}
\begin{defn}
Let $S = \{a_{1}, \ldots, a_{k}\} \subset \R$; $S$ is said to be
{\it algebraically independent over $\Q$} if $f(a_{1}, \ldots, a_{k}) \neq 0$ for all non-trivial $f \in \Q[t_{1}, \ldots, t_{k}]$. $S$ is called {\it algebraically dependent} if it is not algebraically independent.
\end{defn}
Algebraically independent numbers over $\Q$ are ``almost everywhere'' in the
euclidean space:
\begin{lem} \label{thm:measure_algebra}
Let $M > 0$, $n \in \N^{+}$. Consider the uniform measure
(distribution) over $[-M, M]^{n}$; then the measure of the set
$\mathcal{S} = \{(a_{i})_{1 \leq i\leq n} \in [-M, M]^{n}:
\{a_{i}\} \text{ is algebraically independent over } \Q\}$ is $1$.
\end{lem}
By Lemma~\ref{thm:measure_algebra}, we know that if we sample a (real)
matrix uniformly at random, with probability $1$, its entries form an algebraically independent
set over $\Q$. To prove Theorem~\ref{thm:PSDD}, we prove the following intermediate result:
\begin{lem} \label{thm:PSDD_algebra}
Let $B \in \R^{n \times n}$ such that $\{B_{ij}\}$ forms a set of algebraically
independent numbers over $\Q$. Then the size of any PSDD that computes the
L-ensemble with kernel $L = B^{T}B$ is at least $2^{n - 1}$.
\end{lem}
It is clear that Theorem~\ref{thm:PSDD} immediately follows from Lemma
\ref{thm:measure_algebra} and Lemma \ref{thm:PSDD_algebra}, and we briefly
sketch the proof for Lemma~\ref{thm:PSDD_algebra} here.
The proof for Lemma~\ref{thm:PSDD_algebra} relies on the following proposition 
due to \citet{shen2016tractable}, which connects
the size of a PSDD to the number of distinct conditional probability
distributions that it must encode:
\begin{prop}[\citet{shen2016tractable}] \label{prop:PSDD_conditional}
For a PSDD over the variables $X_1, \cdots, X_n$, there exists
a variable $X_{q}$ such that for each $\mathbf{x} \in \{0, 1\}^{n - 1}$,
there is a PSDD node that represents the conditional distribution
$\Pr({X_{q} \given (X_{j})_{j \neq q} = \mathbf{x}})$. In particular,
if $\Pr({X_{q} = 1\given (X_{j})_{j \neq q} = \mathbf{x_1}})
\neq \Pr({X_{q} = 1\given (X_{j})_{j \neq q} = \mathbf{x_2}})$,
then the PSDD nodes that represent them must be distinct.
\end{prop}
By Proposition~\ref{prop:PSDD_conditional}, we know that if we have a
probability distribution such that for all $q \in \{1, \cdots, n\}$, $\Pr(X_{q}
= 1| (X_{j})_{j \neq q} = \mathbf{x})$ is different for all $\mbf{x} \in \{0,
1\}^{n - 1}$, then any PSDD that represents this distribution will have at least
$2^{n - 1}$ nodes. In particular, to prove Lemma~\ref{prop:PSDD_conditional},
we only need to show that the L-ensemble in
Lemma~\ref{thm:PSDD_algebra} satisfies this property; that is,
Lemma~\ref{thm:PSDD_algebra} follows from
Proposition~\ref{prop:PSDD_conditional} and the following result:
\begin{prop} \label{lem:DPP_conditional}
Let $\mathcal{Y} = \{1, \cdots, n\}$ be the ground set.
Consider the L-ensemble $P_{L}$ over $X_1, \cdots, X_n$ with kernel $L = B^{T}B$,
where $\{B_{ij}\}$ is a set of algebraically independent numbers over $\Q$.
Let $\mbf{Y}$ be a random subset drawn from $P_{L}$.
Then $\forall q \in \{1, \cdots, n\}$, the following two equivalent
statements hold:
\begin{enumerate}[noitemsep, topsep=0pt]
\item $P_{L}(X_{q} = 1 \given (X_{j})_{j \neq q} = \mathbf{x})$ is different for
all $\mathbf{x} \in \{0, 1\}^{n-1}$.
\item $P_{L}(\mbf{Y} = A^{in} \cup \{q\} \given
    A^{in} \subset \mbf{Y}, A^{out} \cap \mbf{Y} = \emptyset)$
is different for all disjoint union $\mcal{Y} = A^{in} \cup A^{out} \cup \{q\}$.
\end{enumerate}
\end{prop}

Instead of presenting the complete proof for Proposition~\ref{lem:DPP_conditional},
which is given in the appendix, we leave a short remark to conclude this subsection.
The key point to our proof for Lemma~\ref{thm:PSDD_algebra} is that $f_{A,B,q}$, 
the polynomials defined in Proposition~\ref{prop:polynomial},
do not evaluate to $0$ at point $(B_{ij})$ when the set $\{B_{ij}\}$ is algebraically
independent over $\Q$.
Hence by a continuity argument, it can be shown that
as long as a point $(P_{ij})$ is close enough to $(B_{ij})$,
the L-ensemble with kernel $P^{T}P$ cannot be tractably represented by 
PSDDs. As rational points are dense, we can strengthen Theorem~\ref{thm:PSDD} 
by claiming that besides algebraically independent numbers,
``many'' L-ensembles with
\emph{only rational entries} cannot be represented by polynomial-size PSDDs either.

\subsection{DETERMINISTIC PCs CANNOT TRACTABLY REPRESENT DPPs}

In this section we prove that in general, DPPs cannot be tractably
represented by deterministic PCs.

\begin{thm} \label{thm:deterministic_PC}
Let $\mathcal{Y} = \{1, \cdots, n\}$ be the ground set.
For all $n \geq 2$, there exists an L-ensemble
$P_{n}$ over the subsets of $\mcal{Y}$ such that the number of the parameters in any
deterministic AC that represents $P_{n}$ is at least $2^{n} - 2$.
\end{thm}

To prove Theorem \ref{thm:deterministic_PC}, we first consider the constraints that determinism put on PCs:
\begin{lem} \label{lem:deterministic_PC}
Let $\mathscr{A}$ be a deterministic PC over variables $\mbf{X}$;
let $\theta_{1}, \cdots, \theta_{m} \in \R$ be the parameters in $\mathscr{A}$.
Let $f$ be the function that $\mathscr{A}$ computes.
Then, for any instantiation $\mbf{X} = \mbf{x}$,
$f(\mathbf{x}) = 0$ or $\prod_{1 \leq i\leq m} \theta_{i}^{a_{i}}$ for
some $a_{i} \in \N$.
\end{lem}
Before presenting the formal proof for Theorem~\ref{thm:deterministic_PC},
we first sketch the intuition here.
For a deterministic PC over $n$ variables, there are $2^{n}$ possible instantiations
to its variables, which yields $2^{n}$ probabilities denoted by $p_0, \cdots, p_{2^{n} - 1}$.
By Lemma \ref{lem:deterministic_PC}, we know that every $p_i$ is either $0$ or some
products of the parameters in the PC. 
If the number of parameters is less than $2^{n}$, then the $p_i$s are related to each
other by multiplication; that is, if $p_{i} \neq 0$, then
$p_{i}^{r} = \prod_{j \neq i} p_{j}^{r_{j}}$ for some $r, r_{j} \in \N$.
Now the proof for Theorem~\ref{thm:deterministic_PC}
reduces to constructing a class of L-ensembles such that for all
$i$ the equalities $p_{i}^{r} = \prod_{j \neq i} p_{j}^{r_{j}}$ \emph{do not hold},
which we formalize as condition 2 in Lemma \ref{lem:construction}, and this
is the key property that we are aiming for.
\begin{lem} \label{lem:construction}
There exists a sequence (indexed by $n$)
of matrices $L_{n} \in \R^{n \times n}$ where each $L_{n}$ is of the form
$$
L_{n} =
\begin{bmatrix}
    d_{1} & 1 & \dots & 1 \\
    1 & d_{2} &  & 1 \\
    \vdots & & \ddots & \vdots \\
    1 & \ldots & & d_{n} \\
\end{bmatrix}
$$
and satisfies the following conditions
\begin{enumerate}[noitemsep,topsep=0pt]
\item $\det((L_{n})_{A}) > 1$, $\forall A \subset [n]$ s.t. $A \neq \emptyset$, which implies that
    $L_{n}$ defines an L-ensemble.
\item $(\det((L_{n})_{A}))^{r} \neq \prod_{B \subset [n], B \neq A} (\det((L_{n})_{B}))^{r_{B}}$,
    $\forall A \subset [n]$, $r, r_{B} \in \N$ s.t. $A \neq \emptyset$, $r \neq 0$.
\end{enumerate}
\end{lem}
Now we proceed to prove Theorem \ref{thm:deterministic_PC}.
\begin{proof}[Proof for Theorem \ref{thm:deterministic_PC}]
Let $L \in \R^{n \times n}$ be the $n$th kernel matrix that we constructed in
Lemma \ref{lem:construction}; let $P$ be the DPP defined by $L$. Let $\mathscr{A}$
be a deterministic PC with parameters $\{\theta_{i}\}_{1 \leq i \leq m}$. Assume that
$\mathscr{A}$ computes the unnormalized $P$; that is, assume
$\mathscr{A}(\mbf{x}) = \det (L_{\mbf{x}})$ for all
$\mbf{x} \in \{0, 1\}^{n}$. Suppose (towards a contradiction) that $m < 2^{n} - 1$.
We write the set $\{\det (L_{A}): A \subset [n]\}$ as $\{p_{i}\}_{0 \leq i < 2^{n}}$
where $p_{0} = det(L_{\emptyset}) = 1$. By Lemma~\ref{lem:construction}, $p_{i} > 1$
$\forall i \neq 0$. By Lemma~\ref{lem:deterministic_PC}, we have:
$
p_{i} = \prod_{1 \leq j \leq m} \theta_{j}^{v_{ij}}
$
for some $v_{ij} \in \N$, $0 < i < 2^{n}$.

Put $v_{i} = [v_{i1}, \ldots, v_{im}]^{T} \in \Q^{m}$, $\forall 0 < i < 2^{n}$. Let
$S = \{v_{i}\}_{0 < i < 2^{n}}$. Note that $\forall 0 < i < 2^{n}$,
$p_{i} > 1 \implies v_{i} \neq 0$. Since $\Q^{m}$ is an $m$-dimensional vector space
over $\Q$, $|S| = 2^{n} - 1 > m \implies$ $S$ is linearly dependent. Without loss
of generality, $v_{1} = \sum_{2 \leq k < 2^{n}} q_{k}v_{k}$ for some $q_{k} \in \Q$,
where some $q_{k}$ are non-zero. Hence, $v_{1j} = \sum_{2 \leq k < 2^{n}}q_{k}v_{kj}$;
then,
\begin{align*}
p_{1} =& \prod_{1 \leq j \leq m} \theta_{j}^{v_{1j}} 
      = \prod_{1 \leq j \leq m} \theta_{j}^{\sum_{2 \leq k < 2^{n}}q_{k}v_{kj}} \\
      =& \prod_{1 \leq j \leq m} \prod_{2 \leq k < 2^{n}} \theta_{j}^{q_{k}v_{kj}} 
      = \prod_{2 \leq k < 2^{n}} (\prod_{1 \leq j \leq m} \theta_{j}^{v_{kj}})^{q_{k}} \\
      =& \prod_{2 \leq k < 2^{n}} p_{k}^{q_{k}}
\end{align*}

Write $q_{k} = \frac{r_{k}}{s_{k}}$; put $l = \text{lcm}\{s_{k}: q_{k} \neq 0\}$. $l$
is well-defined and clearly non-zero. Then,
$$
p_{1}^{l} = \prod_{2 \leq k < 2^{n}} p_{k}^{\frac{r_{k}}{s_{k}}l}
$$
That is, for some natural numbers $\{r_{B}\}_{B \subset [n]}$, $A \subset [n]$,
$$
(\det(L_{A}))^{l} = \prod_{B \subset [n], B \neq A} (\det (L_{B}))^{r_{B}}
$$
where $l \neq 0$; contradicting the choice of $L$.
Hence $m \geq 2^{n} - 1$ must follow.
Now assume that $\mathscr{A}$ computes $P$ (i.e. normalized); suppose $m < 2^{n} - 2$;
then we can easily obtain a deterministic AC $\mathscr{A}^{\prime}$ that has
$m + 1 < 2^{n} - 1$ parameters and computes the unormalized $P$; by contradiction,
$m \geq 2^{n} - 2$ must follow.
\end{proof}

\subsection{NO TRACTABLE SPNs FOR DPPs}
In this section, we prove that DPPs in general cannot be tractably represented
by decomposable PCs with no negative parameters, often called sum-product networks (SPNs)~\citep{poon2011sum}. In particular,
the uniform distribution over spanning trees on complete graphs is a DPP that has no 
polynomial-size SPN representation.
\begin{thm} \label{thm:decomposable_PC}
For all $n \geq 20$, let $\mathcal{Y} = \{1, 2, \cdots, {{n}\choose{2}}\}$ be the ground set;
then there exists a DPP over the subsets of $\mcal{Y}$ such that the size of
any smooth SPN that represents the DPP is
at least $2^{n / 30240}$.
\end{thm}
An SPN is \emph{smooth} if for each $\oplus$ gate, all of its children mention the same
set of variables~\citep{darwiche2000tractable}. 
Decomposable circuits  can be made smooth efficiently~\citep{shih2019smoothing}.
Hence, Theorem \ref{thm:decomposable_PC} immediately implies
that in general, DPPs cannot be tractably represented by~SPNs.

Now we give the connection to distributions on spanning trees.
Let $K_{n}$ be the complete graph on $n$ vertices. For each edge $e_{ij}$ with
$1 \leq i < j \leq n$, we associate an indicator variable $X_{ij}$. By Cayley's formula
there are $n^{n - 2}$ spanning trees on $K_{n}$~\citep{cayley1889theorem}.
Hence, we can define a
distribution on $\mbf{X} = (X_{ij})_{1 \leq i < j \leq n}$ by setting
$\Pr(\mbf{X} = \mbf{x}) = 1 / n^{n - 2}$ if the set $\{e_{ij}: x_{ij} = 1\}$
forms a spanning tree on $K_{n}$ and $0$ otherwise. We call it the {\it uniform distribution
over spanning trees} on the complete graph. Note that the distribution is defined on
${n}\choose{2}$ variables. The uniform distribution over spanning 
trees {\it is a DPP}~\citep{snell1995topics}:
\begin{thm} \label{thm:ST_DPP}
The uniform distribution over spanning trees on the complete graph $K_{n}$ is
a DPP, where the marginal kernel is given by:
$$
K(e_{ij}, e_{lk}) =
\begin{cases}
\frac{2}{n} & \text{if } i = k, j = l \\
\frac{1}{n} & \text{if } i = k, j \neq l \text{ or } i \neq k, j = l \\
-\frac{1}{n} & \text{if } j = k \text{ or } i = l \\
0 & \text{otherwise} \\
\end{cases}
$$
$\text{where } 1 \leq i < j \leq n \text{, } 1 \leq l < k \leq n$
\end{thm}

\begin{figure}[]
\centering
\includegraphics[width=0.85\columnwidth]{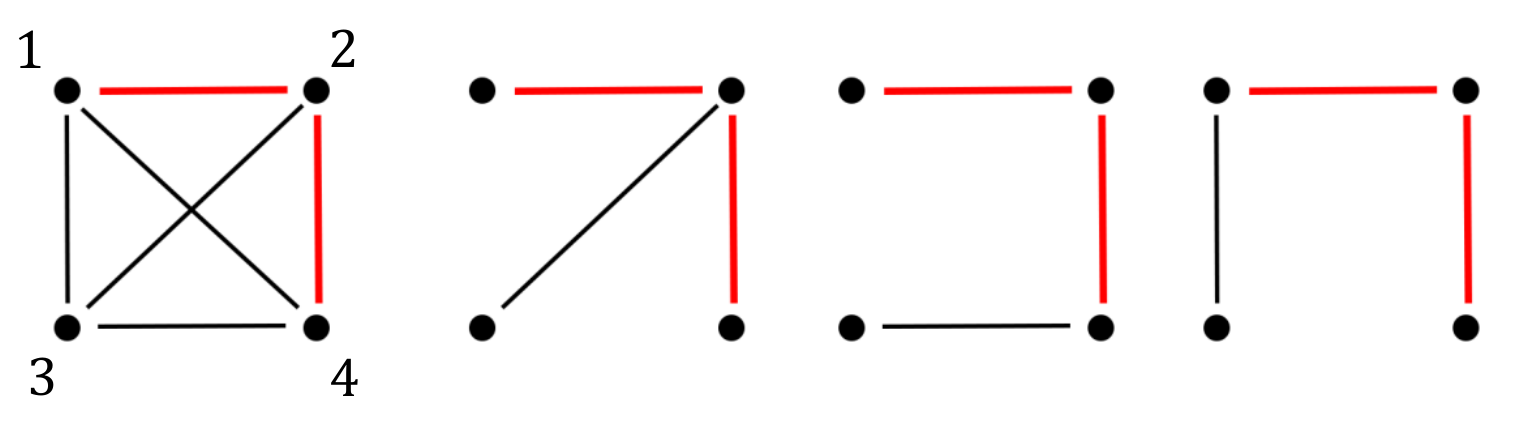}
\caption{The complete graph on $4$ vertices $K_{4}$ and the $3$ spanning
trees that contain edges $e_{12}$ and $e_{24}$}
\label{spanning_tree}
\end{figure}

To get a concrete understanding of the theorem, consider the following example
where $n = 4$.
Here for the complete graph with $4$ vertices, the marginal kernel $K$ is
{\footnotesize
$$
\begin{blockarray}{ccccccc}
    & e_{12} & e_{13} & e_{14} & e_{23} & e_{24} & e_{34} \\
\begin{block}{c[cccccc]}
  e_{12} & 1/2 & 1/4 & 1/4 & -1/4 & -1/4 & 0 \\
  e_{13} & 1/4 & 1/2 & 1/4 & 1/4  & 0 & -1/4 \\
  e_{14} & 1/4 & 1/4 & 1/2 & 0  & 1/4 & 1/4 \\
  e_{23} & -1/4 & 1/4 & 0 & 1/2 & 1/4 & -1/4 \\
  e_{24} & -1/4 & 0 & 1/4 & 1/4 & 1/2 & 1/4 \\
  e_{34} & 0 & -1/4 & 1/4 & -1/4 & 1/4 & 1/2 \\
\end{block}
\end{blockarray}
$$}
To compute the number of spanning trees that contain edges $e_{12}$ and
$e_{24}$, we compute the determinant of the submatrix indexed by $e_{12}$ and $e_{24}$:
$$\Pr(X_{12} = 1, X_{24} = 1) =
\begin{vmatrix}
    1/2 & -1/4 \\
    -1/4 & 1/2 \\
\end{vmatrix}
= 3/16
$$
The total number of spanning trees on $K_{4}$ is $4^{4 - 2} = 16$;
hence, the number of spanning trees that contain $e_{12}$ and $e_{24}$
is given by $3 / 16 \times 16 = 3$, illustrated by Figure~\ref{spanning_tree}.

Now we present the second result: the uniform distributions over spanning
trees cannot be tractably represented by SPNs.\footnote{Note that Theorem~\ref{thm:decomposable_PC} only holds for DPPs, because the uniform distribution over spanning trees is {\it not} an
 L-ensemble.}
\begin{thm}[\citet{martens2014expressive}] \label{thm:ST_SPN}
Let $P_{n}$ be the uniform distribution over spanning trees on $K_{n}$.
For $n \geq 20$, the size of any smooth SPN that represents $P_{n}$ is at least
$2^{n / 30240}$.
\end{thm}
Theorem \ref{thm:decomposable_PC} immediately follows
from Theorems \ref{thm:ST_DPP} and \ref{thm:ST_SPN}.
\section{PERSPECTIVES}
\label{sec:discussion}
So far we have proved three negative results on the problem of representing
DPPs by PCs and we briefly summarize them as follows:
\begin{enumerate}[noitemsep, leftmargin=*, topsep=0pt]
\item For an L-ensemble with kernel $L = B^{T}B$, where $B \in \R^{n \times n}$
is randomly generated, with probability $1$, the L-ensemble cannot be tractably
represented by PSDDs.
\item There exists a class of L-ensembles that cannot be tractably represented
by deterministic PCs with (possibly) negative parameters.
\item There exists a class of DPPs that cannot be tractably represented
by SPNs with non-negative parameters.
\end{enumerate}

This paper closes many avenues for finding a unified tractable probabilistic model.
In this section we discuss some promising avenues that remain open. In particular,
we propose the problem of representing certain \emph{subclasses} of DPPs by
polynomial-size PCs.

\subsection{A SIMPLE CASE STUDY}

As shown in Figure~\ref{fig:motivation}, there is one degenerate class of DPPs
that we can tractably represent
by deterministic and decomposable PCs: the fully factorized distributions.
This subclass of DPP is uninteresting
because it expresses no dependence among the variables.
Inspired by the Matrix Determinant Lemma~\citep{harville1998matrix},
which is Lemma~\ref{lem:matrix_determinant} in the appendix,
we consider a
class of non-trivial L-ensembles that exhibits negative dependence among the variables.
In particular, for the ground set $\mathcal{Y} = \{1, \cdots, n\}$,
we consider \emph{rank-$1$ perturbation} (R1P) L-ensembles with kernels of the form $L = D + \lambda u u^{T}$, where
$D = \text{diag}(d_{1}, \cdots, d_{n})$, $d_{i} \geq 0, \lambda \geq 0$ and
$u = (u_1, \cdots, u_n)^T \in \R^{n}$ a normal vector. 
It is clear that for kernel $L$, the off-diagonal entries
$L_{ij} = \lambda u_{i} u_{j}$ can be set to non-zero.
Hence, this special class of L-ensembles \emph{does} exhibit 
negative dependence among
variables to some extent, which is one of the key properties that makes DPPs
a unique class of TPMs. Now we ask two questions: (1) can R1P L-ensembles be represented tractably by structured PCs; (2) can other more complex DPPs be represented as mixtures of R1Ps.

Let $\mbf{Y}$ be a subset drawn from the L-ensemble with kernel $L$;
let $A \subseteq \mathcal{Y}$, then by
the matrix determinant lemma:
$$
\Pr(\mbf{Y} = A)
= \frac{1}{Z} (\prod_{i \in A} d_{i} + \sum_{i \in A} {u_{i}}^2
                                          \prod_{j \in A, j \neq i} d_{j}),
$$
where the normalizing constant $Z = \det(D + \lambda u u^{T} + I)$.
This formula immediately gives us a decomposable PC that represents $L$,
which is shown in Figure~\ref{fig:dPC_L} in the appendix; we denote the circuit by $\mathcal{C}$. Thus we have an example of an interesting DPP represented
as a polynomial-size structured PC, answering (1) in the positive.

Now we explore question (2) empirically on a small example.
In a preliminary study, we approximate an L-ensemble
$P_L$ with kernel $L = B^{T}B$, where $B \in \R^{K \times N}$ is randomly generated. 
To do this, we take a weighted sum of $m$ instances of $\mathcal{C}$
to approximate $P_L$, which we call a \emph{R1P mixture}. For different values of $m$, we compare our mixture model
against the weighted sum of $2m$ instances of the fully-factorized
distribution (the baseline); this makes the comparison fair in the sense that both models have $2mN$
parameters. 
For both models,
the parameters are learned by stochastic gradient descent to minimize the
KL-divergence between the model and $P_L$. 
We initialize the parameters randomly
and record the best results (minimum KL-divergence) from 20 instances of
random restarts.

Table~\ref{fig:experiment} gives the results of this preliminary experiment. It shows that this is a potential promising avenue for approximating DPPs with structured circuits, motivating future empirical and theoretical work. In particular, we see that
the ratio between the KL-divergence for the baseline and the R1P mixture
increases as the number
of components in the mixtures increases.

\begin{table}
  \centering
{\footnotesize 
  \begin{tabular} {cccc}
    \toprule
$m$ & Fully-Factorized KL & R1P Mixture KL & Ratio \\
    \midrule
1  & 0.23406 & {\bf 0.23240} & 1.00 \\ 
2  & 0.14948 & {\bf 0.14778} & 1.01 \\
8  & 0.03963 & {\bf 0.03690} & 1.07 \\
16 & 0.01373 & {\bf 0.01077} & 1.27 \\
24 & 0.00554 & {\bf 0.00381} & 1.45 \\
32 & 0.00264 & {\bf 0.00162} & 1.62 \\
40 & 0.00125 & {\bf 0.00062} & 2.02 \\
47 & 0.00054 & {\bf 0.00027} & 1.99 \\
    \bottomrule
\end{tabular}
}
\caption{Results for the motivating experiment on approximating an L-ensemble
over $N = 10$ variables. The second and third column gives the KL divergence between 
the L-ensemble and the approximating family. The fourth column highlights the trend  by giving the ratio between the 
second and third column.
}
\label{fig:experiment}
\end{table}

\subsection{DISCUSSION AND CONCLUSION}
We conclude by highlighting future directions and contextualizing our 
preliminary experimental results.
Though we have proved that PSDDs, deterministic PCs with (possibly) negative
parameters and SPNs cannot tractably represent DPPs in general, some
problems on the existence of polynomial-size PCs that represent DPPs
are left open:
\begin{enumerate}[noitemsep, topsep=0pt]
\item Can the uniform distribution over spanning trees on complete graphs
be tractably represented by decomposable PCs with negative parameters?
\item Can L-ensembles be tractably represented by decomposable PCs (with or
without negative parameters)?
\end{enumerate}
These two open problems are important not only because they are the missing pieces
for a complete answer to the problem of representing DPPs by polynomial-size PCs, but also
because their answers will deepen our understanding about more general questions.
An answer to question~1 will help us better understand the
expressive power that PCs gain by allowing negative parameters.

If the answer to question 2 is positive; then we immediately have a class
of decomposable PCs that can tractably represent L-ensembles, which implies
that there exists a class of decomposable PCs that represents probability distributions
with global negative dependence, which, as mentioned before, is intractable
for some graphical models. Besides, starting from polynomial-size 
decomposable PCs that represents L-ensembles, we can obtain decomposable PCs
that exhibit global negative dependence and at the same time allow for greater
flexibility than L-ensembles do. If the answer to question 2 is negative, it will still deepen
our understanding about the expressiveness of polynomial-size PCs.

To conclude, this work identifies and makes progress towards closing 
a gap between two well-known families of TPMs. We hope that this effort
motivates future work in extending our understanding of the expressiveness of probabilistic
circuits and motivates further efforts in combining and comparing different
classes of TPMs.

\subsubsection*{Acknowledgements}
We thank Andreas Krause, Arthur Choi, Brendan Juba, Pasha Khosravi, Anji Liu, Tao Meng,
Antonio Vergari and Zhe Zeng
for helpful feedback and discussion. In particular, we thank
Richard Elman for detailed and insightful discussion on algebra.
This work is partially supported by NSF grants \#IIS-1943641, \#IIS-1633857,
\#CCF-1837129, DARPA XAI grant \#N66001-17-2-4032, Sloan and UCLA Samueli Fellowships, 
and gifts from Intel and Facebook Research.

\bibliographystyle{plainnat}
\bibliography{ac_dpp}

\onecolumn

\appendix
\section{FIGURES}

\begin{figure}[ht]
\vskip 0.2in
\begin{center}
\centerline{\includegraphics[width=0.7\textwidth]{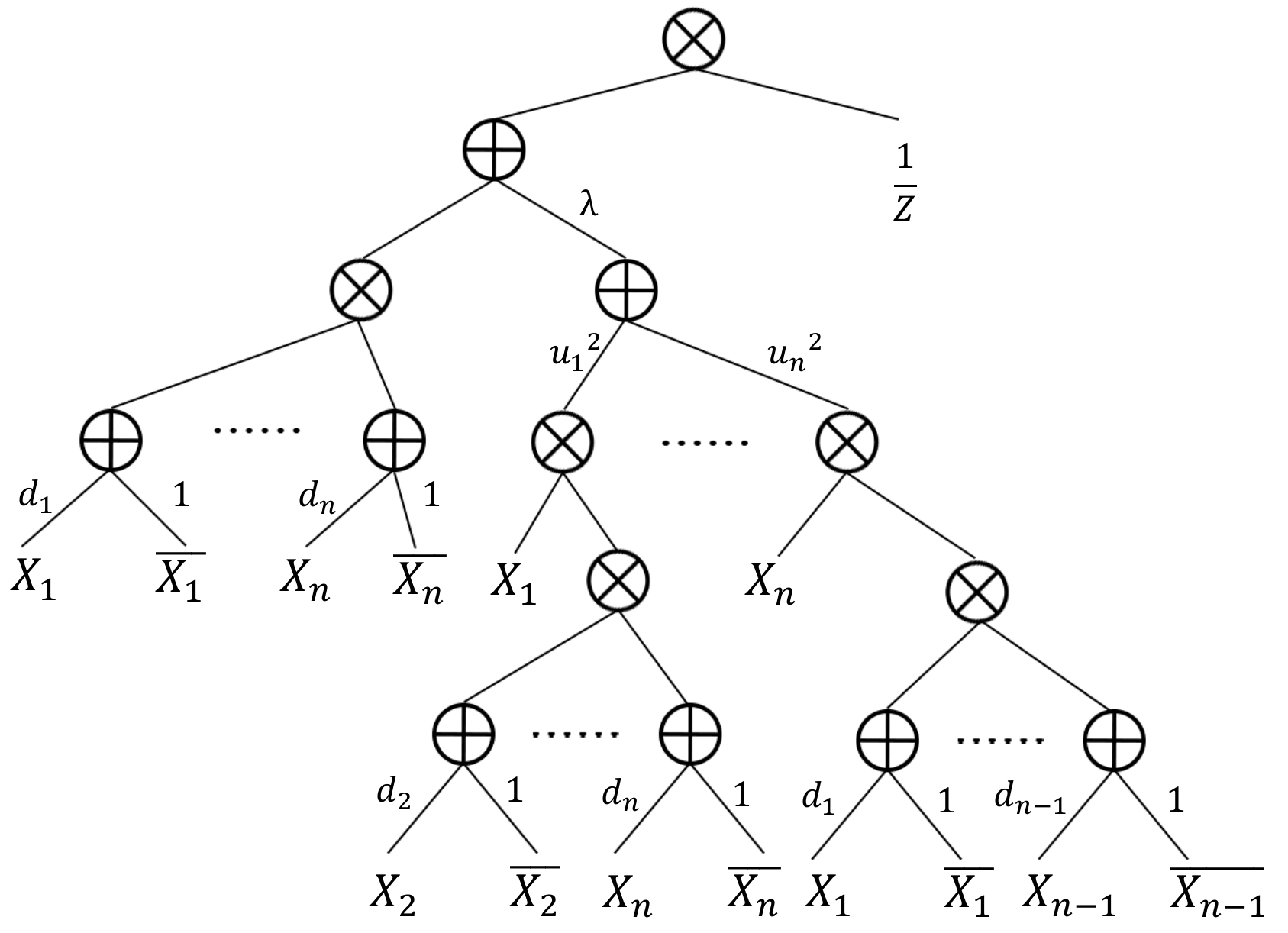}}
\caption{A polynomial-size decomposable PC that represents L-ensembles
with kernels of the form $L = D + \lambda u u^{T}$.}
\label{fig:dPC_L}
\end{center}
\end{figure}

\section{LEMMAS AND PROOFS}
\begin{lem} \label{lem:measure_roots}
Let $M > 0$; let $n \geq 1$. Let $\mu_{n}$ be the uniform measure (normalized
Lebesgue measure) over $[-M, M]^{n}$. Let $0 \neq f \in \R[t_{1}, \cdots, t_{n}]$.
Let $S_{f}$ be the set of roots for $f$ in $[-M, M]^{n}$, then $\mu_{n}(S_f) = 0$
\end{lem}
\begin{proof}[Proof for Lemma \ref{lem:measure_roots}]
We prove the claim by induction on $n$. If $n = 1$, then a non-trivial polynomial
with a single variable has finitely many roots in $\R$; done.
Now assume the claim holds for $n$.
Let $0 \neq f \in \R[t_{1}, \cdots, t_{n + 1}]$. Note that we can also view
$f$ as a single-variable polynomial with respect to $t_{n + 1}$; that is,
$0\neq f \in (\R[t_{1}, \cdots, t_{n}])[t_{n + 1}]$.
Since $\R[t_{1}, \cdots, t_{n}]$ is an {\it integral domain}, there exist
finitely many $r \in \R[t_{1}, \cdots, t_{n}]$ s.t.
$f(t_{1}, \cdots, t_{n}, r) = 0$;
in particular, there exist finitely many $r \in [-M, M]$ s.t. $f(t_1, \cdots, t_n, r)$
is the trivial polynomial with respect to $t_1, \cdots, t_n$,
denote them by $r_1, \cdots, r_k$.
Now let $x \in [-M, M]$, put $f_{x} = f(t_1, \cdots, t_n, x) \in \R[t_1, \cdots, t_n]$;
note that, by the induction hypothesis $\mu_{n}(S_{f_x}) = 0$
if and only if $f_x \neq 0$ if and only if $x \neq r_{i}$ $\forall 1 \leq i \leq k$.
Hence, since $S_f = \bigcup_{x \in [-M, M]} S_{f_x} \times \{x\}$, then
by Fubini's Theorem, we have
\begin{align*}
  &\mu_{n+1}(S_f)\\
= &\int_{x \in [-M, M]} \mu_{n}(S_{f_x}) \\
= &\int_{x \in [-M, M] - \{r_i\}} \mu_{n}(S_{f_x})
  + \int_{x \in \{r_i\}} \mu_{n}(S_{f_x}) \\
= &\int_{x \in [-M, M] - \{r_i\}} 0
  + \int_{x \in \{r_i\}} \mu_{n}(S_{f_x}) \\
= &0
\end{align*}
\end{proof}

\begin{proof}[Proof for Lemma \ref{thm:measure_algebra}]
We prove that the measure of the complement of $\mathcal{S}$ is $0$.
That is, by the definition of algebraically independent numbers, we
prove that the set $\mathcal{S}^{C} = \{(a_{i}) \in  [-M, M]^{n}:
f(a_{1}, \cdots, a_{n}) = 0 \text{ for some } 0\neq f \in \Q[t_1, \cdots, t_n]\}$
has measure 0. By Lemma \ref{lem:measure_roots}, for each
$0 \neq f \in \Q[t_1, \cdots, t_n]$, the set of roots for $f$ in $[-M, M]^{n}$
has measure $0$; since the set $\Q[t_1, \cdots, t_n]$ is countable, we
can conclude that
$\mathcal{S}^{C} = \bigcup_{0 \neq f \in \Q[t_1, \cdots, t_n]}
\{\text{set of roots for } f \text{ in } [-M, M]^n\}$ has measure $0$.
\end{proof}

Before we prove Proposition~\ref{lem:DPP_conditional}, we first prove the 
following result.

\begin{prop} \label{prop:polynomial}
Let $n \geq 2$. Let $\td{D} = [t_{ij}]_{1 \leq i,j \leq n}$ be a $n \times n$ matrix
where the entries $t_{ij}$ are different indeterminates. Put $\td{L} = {\td{D}}^{T}\td{D}$,
then the following polynomial in $\Q[(t_{ij})_{1 \leq i,j \leq n}]$
\begin{equation}
\label{eq:polynomial}
f_{A, B, q} = \det(\td{L}_{A})\det(\td{L}_{B\cup\{q\}})
            - \det(\td{L}_{B})\det(\td{L}_{A\cup\{q\}})
\end{equation}
is non-trivial $\forall A, B \subset \{1, \cdots, n\}, q \in \{1, \cdots, n\}$
such that $A \neq B$, $q \notin A$, $q \notin B$.
\end{prop}

\begin{proof}[Proof for Proposition \ref{prop:polynomial}]
Fix $q \in [n]$. Let $A, B$ be subsets of $[n]$. Assume that $A \neq B$, $q \notin A$
and $q \notin B$. It is immediate from definition that
$f_{A, B, q} \in \Q[(t_{ij})_{1 \leq i,j \leq n}]$.
To show that $f_{A, B, q} \neq 0$, we just need to find one point
$\mathbf{x} \in R^{n \times n}$ such that $f_{A, B, q}|_{\mathbf{x}} \neq 0$; consider
the following evaluation at $\mathbf{x}$:
$$\td{D}|_{\mathbf{x}} =
\begin{bmatrix}
    1 & 0 & \dots & (\sqrt2)^{1} & \dots & 0 \\
    0 & 1 & \dots & (\sqrt2)^{2} & \dots & 0 \\
    \vdots & \vdots & \ddots & \vdots & & \vdots \\
    0 & 0 & \dots & (\sqrt2)^{q} & \dots & 0 \\
    \vdots & \vdots &  & \vdots & \ddots & \vdots \\
    0 & 0 & \dots & (\sqrt2)^{n} & \dots & 1 \\
\end{bmatrix}
\text{  that is, }
\begin{cases}
    t_{ii} = 1 & \forall i \neq q\\
    t_{qi} = (\sqrt2)^{i} \\
    t_{ij} = 0 & otherwise \\
\end{cases}
$$
By computation,
$$L:=\td{L}|_{\mathbf{x}} =
\begin{bmatrix}
    1 & 0 & \dots & (\sqrt2)^{1} & \dots & 0 \\
    0 & 1 & \dots & (\sqrt2)^{2} & \dots & 0 \\
    \vdots & \vdots & \ddots & \vdots & & \vdots \\
    (\sqrt2)^{1} & (\sqrt2)^{2} & \dots & s & \dots & (\sqrt2)^{n} \\
    \vdots & \vdots &  & \vdots & \ddots & \vdots \\
    0 & 0 & \dots & (\sqrt2)^{n} & \dots & 1 \\
\end{bmatrix}
\text{\quad where }
L_{ij} =
\begin{cases}
     \delta_{ij} & i \neq q, j \neq q\\
    (\sqrt2)^{i} & i = q, j \neq q\\
    (\sqrt2)^{j} & i \neq q, j = q\\
    s := 2^{n + 1} - 2 & i = q, j = q\\
\end{cases}
$$
First, note that $L_{C} = I$ for all $C \subset [n]-\{q\}$; hence
$\det(L_{A}) = \det(L_{B}) = 1$ and it follows from Equation~\ref{eq:polynomial} that
$$f_{A,B,q}|_{\mathbf{x}} = \det(L_{B \cup \{q\}}) - \det(L_{A \cup \{q\}})$$
Then let's compute $\det(L_{B \cup \{q\}})$. Write $B = \{b_{1}, \dots, b_{k}\}$ where
$k = |B|, b_{1} < b_{2} < \dots< b_{k}$, then:
$$L_{B \cup \{q\}} =
\begin{bmatrix}
    1 & 0 & \dots & (\sqrt2)^{b_{1}} & \dots & 0 \\
    0 & 1 & \dots & (\sqrt2)^{b_{2}} & \dots & 0 \\
    \vdots & \vdots & \ddots & \vdots & & \vdots \\
    (\sqrt2)^{b_{1}} & (\sqrt2)^{b_{2}} & \dots & s & \dots & (\sqrt2)^{b_{k}} \\
    \vdots & \vdots &  & \vdots & \ddots & \vdots \\
    0 & 0 & \dots & (\sqrt2)^{b_{k}} & \dots & 1 \\
\end{bmatrix}
$$
After swapping rows and columns, we have:
$$\det(L_{B \cup \{q\}}) =
\begin{vmatrix}
    s & (\sqrt2)^{b_{1}} & (\sqrt2)^{b_{2}} & \dots & (\sqrt2)^{b_{k}} \\
    (\sqrt2)^{b_{1}} & 1 & 0 & \dots & 0 \\
    (\sqrt2)^{b_{2}} & 0 & 1 & & 0 \\
    \vdots & \vdots & & \ddots &  \\
    (\sqrt2)^{b_{k}} & 0 & 0 &  & 1 \\
\end{vmatrix}
$$
We compute $\det(L_{B \cup \{q\}})$ by Laplace expansion on the first row:
$$\det(L_{B \cup \{q\}}) = s + (-1)^{1}(\sqrt{2})^{b_{1}}M_{12}                             
                             + \dots
                             + (-1)^{k}(\sqrt{2})^{b_{k}}M_{1(k+1)}$$
where $M_{1j}$ denotes the $(1, j)$ minor for $L_{B\cup\{q\}}$.\\
\\
Now we compute $M_{1j}$ for $2 \leq j \leq k + 1$:
$$
M_{1j} =
\begin{vmatrix}
    (\sqrt2)^{b_{1}} & 1 & \dots & 0 & 0 & \dots & 0\\
    \vdots & \ddots & \ddots &  \\
     & & \ddots & 1 & \\
    (\sqrt2)^{b_{j - 1}} & & & 0 & 0 \\
     & & & & 1 & \ddots \\
    \vdots & & & & & \ddots & 0 \\
    (\sqrt2)^{b_{k}} & 0 & \dots & 0 & 0 & \dots & 1 \\
\end{vmatrix}
$$
We can swap the first column with the following column for $j - 2$ times then get
\begin{align*}
M_{1j} &= (-1)^{j-2}
\begin{vmatrix}
    1 & \dots & 0 & (\sqrt2)^{b_{1}} & \dots & 0\\
    \vdots & \ddots & & \vdots & & \\
     & & 1 & \\
    0 & & & (\sqrt2)^{b_{j - 1}} & & & \\
    & & & & \ddots & \\
    0 & & \dots & (\sqrt2)^{b_{k}} & 0 & 1 \\
\end{vmatrix}\\
&=(-1)^{j-2}(\sqrt{2})^{b_{j-1}}
\end{align*}
Hence,
\begin{align*}
\det(L_{B \cup \{q\}}) & = s + \sum_{j=2}^{k+1}(-1)^{j-1}(\sqrt{2})^{b_{j-1}}M_{1j}\\
                       & = s + \sum_{j=2}^{k+1}(-1)^{j-1}(\sqrt{2})^{b_{j-1}}(-1)^{j-2}(\sqrt{2})^{b_{j-1}}\\
                       & = s - \sum_{j=1}^{k}2^{b_{j}}\\
\end{align*}
Similarly, if we write $A = \{a_{1}, \dots, a_{l}\}$, then
$$\det(L_{A \cup \{q\}}) = s - \sum_{j=1}^{l}2^{a_{j}}$$
Thus,
\begin{align*}
f_{A,B,q}|_{\mathbf{x}} & = \det(L_{B \cup \{q\}}) - \det(L_{A \cup \{q\}})\\
                        & = (s - \sum_{j=1}^{k}2^{b_{j}}) - (s - \sum_{j=1}^{l}2^{a_{j}})\\
                        & = \sum_{j=1}^{l}2^{a_{j}} - \sum_{j=1}^{k}2^{b_{j}}\\
\end{align*}
Consider the binary representation of natural numbers; it is then immediate that
$A \neq B \implies {f_{A,B,q}|_{\mathbf{x}} \neq 0} \implies f_{A,B,q} \neq 0$.
\end{proof}

Now we proceed to prove Proposition~\ref{lem:DPP_conditional}.

\begin{proof}[Proof for Proposition~\ref{lem:DPP_conditional}]
Let $L=B^{T}B$, where $B$ is defined as in the statement of Proposition \ref{lem:DPP_conditional}.
First, by construction, $L$ is positive semidefinite. Moreover,
as the entries in $B$ are algebraically independent over $\Q$, it follows from definition
that $\det(B) \neq 0$. Hence, the columns of $B$, denote them by $B_{i}$,
are linearly independent. Thus, $\forall A \subset \mathcal{Y}$,
$L_{A} = {[B_{i}]_{i \in A}}^{T}[B_{i}]_{i \in A} \implies \op{rank}(L_{A}) =
\op{rank}([B_{i}]_{i \in A}) = |A|$; that is, $\det(L_{A}) \neq 0$, $\forall A \subset
\mathcal{Y}$.\\
\\
By formula 2.43 from \citet{MAL-044}, given the disjoint
union $\mathcal{Y}=A^{in} \cup A^{out} \cup \{q\}$,
\begin{align*}
 &P_{L}(\mathbf{X} = A^{in} \cup \{q\} | A^{in} \subset \mbf{X}, A^{out} \cap \mbf{X} = \emptyset)\\
= &\frac{\det(L_{A^{in} \cup \{q\}})}{\det(L_{\ol{A^{out}}} + I_{\ol{A^{in}}})}
= \frac{\det(L_{A^{in} \cup \{q\}})}{\det(L_{A^{in} \cup \{q\}} + I_{\{q\}})}\\
= &\frac{\det(L_{A^{in} \cup \{q\}})}{\det(L_{A^{in} \cup \{q\}}) + \det(L_{A^{in}})}    
= \frac{1}{1 + \frac{\det(L_{A^{in}})}{\det(L_{A^{in} \cup \{q\}})}}
\end{align*}
Let $A^{in}_1, A^{in}_2 \subset \mathcal{Y} - \{q\}$ that $A^{in}_1 \neq A^{in}_2$. Suppose we have
$$\frac{\det{L_{A^{in}_1}}}{\det(L_{A^{in}_1 \cup \{q\}})}
= \frac{\det{L_{A^{in}_2}}}{\det(L_{A^{in}_2 \cup \{q\}})}$$
Then, the polynomial defined in Proposition~\ref{prop:polynomial} evaluates to $0$:
$$f_{A^{in}_1,A^{in}_2,q}\Bigr\rvert_{t_{ij}=B_{ij}}
=\det(L_{A^{in}_1})\det(L_{A^{in}_2 \cup \{q\}}) - \det(L_{A^{in}_2})\det(L_{A^{in}_1 \cup \{q\}})
= 0$$

However, by Proposition \ref{prop:polynomial}, $A^{in}_1 \neq A^{in}_2 \implies f_{A^{in}_1,A^{in}_2,q} \in \Q[t_{ij}]$ is
non-trivial $\implies \{B_{ij}\}_{1 \leq i,j \leq n} = S$ is algebraically \emph{dependent} over
$\Q$; contradicting our choice of $B$. Hence,
$$\frac{\det{L_{A^{in}_1}}}{\det(L_{A^{in}_1 \cup \{q\}})}
\neq \frac{\det{L_{A^{in}_2}}}{\det(L_{A^{in}_2 \cup \{q\}})}$$
must follow; that is,
$P_{L}(\mbf{X} = A^{in} \cup \{q\} \given A^{in} \subset \mbf{X}, A^{out} \cap \mbf{X} = \emptyset)$
must be different for all ${A^{in} \subset \mathcal{Y}}$ whenever
$\mathcal{Y} = A^{in} \cup A^{out} \cup \{q\}$ is a disjoint union.
\end{proof}

\begin{lem}[Matrix Determinant Lemma]\label{lem:matrix_determinant}
Let $A$ be an $n \times n$ matrix; let $u$ and $v$ be $n \times 1$ vectors. Then,
$$\det(A + u v^{T}) = \det(A) +v^{T}\text{adj}(A)u$$
where adj$(A)$ is the adjugate of $A$.
\end{lem}

Before proving Lemma \ref{lem:construction}, we first prove the following
computational result.

\begin{lem}\label{lem:diagonal_determinant}
Let $A \in R^{n \times n}$ be a matrix of the form:
$$A =
\begin{bmatrix}
    d_{1} & 1 & \dots & 1 \\
    1 & d_{2} &  & 1 \\
    \vdots & & \ddots & \vdots \\
    1 & \ldots & & d_{n}\\
\end{bmatrix}
$$
then, $$\det A =
\prod_{1\leq i\leq n} (d_{i} - 1) + \sum_{1\leq i\leq n}\prod_{j \neq i}(d_{j}
- 1)$$
\end{lem}
\begin{proof}[Proof for Lemma \ref{lem:diagonal_determinant}]
Let $\mbf{v} = [1, \cdots, 1]^{T} \in {\R}^{n}$;
put $D = \text{diag}(d_{1}, \ldots, d_{n}) - I \in \R^{n \times n}$.
Then by the Matrix Determinant Lemma,
$\det(A) = \det (D + \mbf{v}\mbf{v}^{T}) = \det (D) + \mbf{v}^{T}\text{adj}(D)\mbf{v}$,
where the adjugate of $D$ is given by
$\text{adj}(D) = \text{diag}(\prod_{i \neq 1}(d_{i} - 1), \ldots, \prod_{i \neq n}(d_{i} - 1))$. Hence $\det A = \prod_{i}(d_{i} - 1) + \sum_{i}\prod_{j \neq i}(d_{j} - 1)$
\end{proof}

Now we proceed to prove Lemma \ref{lem:construction}

\begin{proof}[Proof of Lemma \ref{lem:construction}]
We construct a sequence $L_{n}$ inductively.\\
If $n = 1$, put $L_{1} = [2.4]$, which satisfies the conditions trivially.
Now assume that $n \geq 1$ and we have constructed $L_{n}$; put
$$
L(t) =
\begin{bmatrix}
     & & & 1 \\
     & L_{n} & & \vdots \\
     & & &  1 \\
     1 & \ldots & 1 & t
\end{bmatrix}
$$
where $t$ is an indeterminate. Note that if we find an $x \in \R$ s.t. $L(x)$ satisfies
the conditions then we are done by setting $L_{n + 1} = L(x)$.

First note that by Lemma \ref{lem:diagonal_determinant} and the induction hypothesis,
by setting $x > 2$, $L(x)$ immediately satisfies condition 1; and, it is not hard to
observe that $L(x)$ satisfies condition 2 if $x$ is not a root for any
polynomial in the set
$$
P =
\begin{cases}
\disp f_{A, r, \{r_B, s_B\}_{B \subset [n]}} := (\det L(t)_{A \cup \{n + 1\}})^{r}
    - \prod_{B \subset [n], B \neq A} (\det L(t)_{B \cup \{n + 1\}})^{r_{B}}
\prod_{B \subset [n]} (\det L(t)_{B})^{s_{B}}\\

\disp g_{A, r, \{r_B, s_B\}_{B \subset [n]}} := (\det L(t)_{A})^{r}
    - \prod_{B \subset [n]} (\det L(t)_{B \cup \{n + 1\}})^{r_{B}}
\prod_{B \subset [n], B \neq A} (\det L(t)_{B})^{s_{B}} \\

\disp h_{r, \{r_{B}\}_{B \subset [n]}} :=
    (t + 1)^{r} - \prod_{B \subset [n]}(\det L(t)_{B})^{r_{B}} \\
\end{cases}
$$
where $A \subset [n]$, $r, r_{B}, s_{B} \in \N$ such that $A \neq \emptyset$, $r \neq 0$.
For simplicity we will denote these polynomials by $f$, $g$ and $h$; we first show that
all polynomials in P are non-trivial.\\
\\
Denote the diagonal elements of $L_n$ by $d_{1}, \ldots, d_{n}$;
then by Lemma \ref{lem:diagonal_determinant},
\begin{equation} \label{eq:detLt}
\begin{split}
    &\det (L(t)_{A \cup \{n + 1\}}) \\
    =& (t - 1)\prod_{j \in A}(d_{j} - 1)
    + (t - 1)\sum_{i \in A}\prod_{j \neq i}(d_{j} - 1)
    + \prod_{j \in A}(d_{j} - 1) \\
    =& (t - 1)\left[\prod_{j \in A}(d_{j} - 1) + \sum_{i \in A}\prod_{j \neq i}(d_{j} - 1)\right]
    + \prod_{j \in A}(d_{j} - 1) \\
    =& \det (L(t)_{A})(t - 1) + \prod_{j \in A}(d_{j} - 1)
\end{split}
\end{equation}
Consider $g \in P$; by equation (\ref{eq:detLt}),
\begin{align*}
g = (\det (L(t)_{A}))^{r}
- \prod_{B \subset [n]}\left[\det (L(t)_{B})(t - 1) + \prod_{j \in B}(d_{j} - 1)\right]^{r_{B}}
\prod_{B \subset [n], B \neq A}\det(L(t)_{B})^{s_{B}}
\end{align*}
If $r_{B} = 0$ $\forall B \subset [n]$; then
$g = (\det L(t)_{A})^{r} - \prod_{B \subset [n], B \neq A}(\det L(t)_{B})^{s_{B}} \neq 0$
by the induction hypothesis. If $r_{B} \neq 0$ for some $B \subset [n]$,; then
the coefficient of the highest-degree term in $g$ is given by
$\prod_{B \subset [n]}(\det L(t)_{B})^{r_{B}}
\prod_{B \subset [n], B \neq A}(\det L(t)_{B})^{s_{B}} > 1$ by the induction hypothesis;
hence $g \neq 0$.\\
\\
Consider $f \in P$; again by equation (\ref{eq:detLt}),
\begin{align*}
f = &\left[ \det(L(t)_{A})(t - 1) + \prod_{j \in A}(d_{j} - 1) \right]^{r} \\
- &\prod_{B \subset [n], B \neq A} \left[ \det(L(t)_{B})(t - 1) + \prod_{j\in B} (d_{j} - 1)\right]^{r_{B}}
  \prod_{B \subset[n]} (\det(L(t)_{B})^{s_{B}}
\end{align*}
If $r_{B} = 0$ $\forall B \subset [n]$ s.t. $B \neq A$; then $\text{deg} f = r \geq 1$
$\implies f \neq 0$.\\
If $r_{B} \neq 0$ for some $B \subset [n], B \neq A$; consider the coefficient of the
highest-degree term in f:
\begin{equation} \label{eq:alpha}
\begin{split}
\alpha := &(\det L(t)_{A})^{r} - \prod_{B \subset [n], B \neq A}(\det L(t)_{B})^{r_{B}}
            \prod_{B \subset [n]}(\det L(t)_{B})^{s_{B}}\\
        = &(\det L(t)_{A})^{r}
        - \left[ \prod_{B \subset [n], B \neq A} (\det L(t)_{B})^{r_{B} + s_{B}} \right] (\det L(t)_{A})^{s_{A}}
\end{split}
\end{equation}
We show that $\alpha \neq 0$ case by case.\\
If $r > s_{A} \geq 0$; then, it follows from the induction hypothesis that
\begin{align*}
\alpha = \left[ (\det L(t)_{A})^{r - s_{A}}
- \prod_{B \subset [n], B \neq A} (\det L(t)_{B})^{r_{B} + s_{B}} \right]
(\det L(t))^{s_{A}} \neq 0
\end{align*}
If $s_{A} \geq r$; by the induction hypothesis $\det L(t)_{A} > 1$, hence
$(\det L(t)_{A})^{s_{A}} \geq (\det L(t)_{A})^{r}$. Besides, by the induction
hypothesis and the assumption that $r_{B} > 0$ for some $B \subset [n], B \neq A$, we have $\prod_{B \subset [n], B \neq A} (\det L(t)_{B})^{r_{B} + s_{B}} > 1$; it then follows from equation (\ref{eq:alpha}) that $\alpha < 0$.

Thus $f \neq 0$.

It is clear that $h \in P$ are non-trivial.

Now we know that $P$ is a set of non-trivial polynomials in $\R[t]$; more importantly,
$P$ is countable by definition. Hence, the set of real roots for polynomials in $P$,
$S := \{x \in \R: h(x) = 0 \text{ for some } h \in P\}$, is countable. Thus, we can pick
$x > 2$ s.t. $x \notin S$; $L(x)$ now satisfies both condition 1 and 2 and we are
done by induction.
\end{proof}

\end{document}